\documentclass[10pt,twocolumn,letterpaper]{article}

\usepackage{cvpr}              

\usepackage{tabularx}
\usepackage{graphics} 
\usepackage{amssymb}  
\usepackage{amsmath} 
\usepackage{comment} 
\usepackage{multirow}
\usepackage{multicol}
\usepackage{booktabs}
\usepackage{array}
\usepackage{textcomp}
\usepackage{color}
\usepackage{bm}

\definecolor{cvprblue}{rgb}{0.21,0.49,0.74}
\usepackage[pagebackref,breaklinks,colorlinks,allcolors=cvprblue]{hyperref}

\def\paperID{0029} 
\def\confName{CVPRW}
\def\confYear{2025}

\title{CE-NPBG: \textit{C}onnectivity \textit{E}nhanced \textit{N}eural \textit{P}oint-\textit{B}ased \textit{G}raphics for Novel View Synthesis in Autonomous Driving Scenes}

\author{Mohammad Altillawi$^{1,2 \footnotemark[1]}$, 
Fengyi Shen$^{2,3}$, 
Liudi Yang$^{2,4}$, 
Sai Manoj Prakhya$^{2}$, 
Ziyuan Liu$^{2}$\\
$^{1}$ {\small Computer Vision Center (CVC), Universitat Autònoma de Barcelona}\\ $^{2}$ \small Intelligent Simulation Innovation Lab, Huawei Munich Research Center\\ $^{3}$ \small Technical University of Munich, $^{4}$ \small University of Freiburg\
}

\begin{document}
\maketitle
\begin{abstract}
Current point-based approaches encounter limitations in scalability and rendering quality when using large 3D point cloud maps because using them directly for novel view synthesis (NVS) leads to degraded visualizations. We identify the primary issue behind these low-quality renderings as a visibility mismatch between geometry and appearance, stemming from using these two modalities together. To address this problem, we present CE-NPBG, a new approach for novel view synthesis (NVS) in large-scale autonomous driving scenes. Our method is a neural point-based technique that leverages two modalities: posed images (cameras) and synchronized raw 3D point clouds (LiDAR). We first employ a connectivity relationship graph between appearance and geometry, which retrieves points from a large 3D point cloud map observed from the current camera perspective and uses them for rendering. By leveraging this connectivity, our method significantly improves rendering quality and enhances run-time and scalability by using only a small subset of points from the large 3D point cloud map. Our approach associates neural descriptors with the points and uses them to synthesize views. To enhance the encoding of these descriptors and elevate rendering quality, we propose a joint adversarial and point rasterization training. During training, we pair an image-synthesizer network with a multi-resolution discriminator. At inference, we decouple them and use the image-synthesizer to generate novel views. We also integrate our proposal into the recent 3D Gaussian Splatting work to highlight its benefits for improved rendering and scalability.
\end{abstract}

\footnotetext[1]{Corresponding author. {email: mohammad.altillawi1@huawei.com}}    
\section{Introduction}
\label{sec:intro}

\begin{figure}[t]

  \centering
    \includegraphics[scale=0.242]{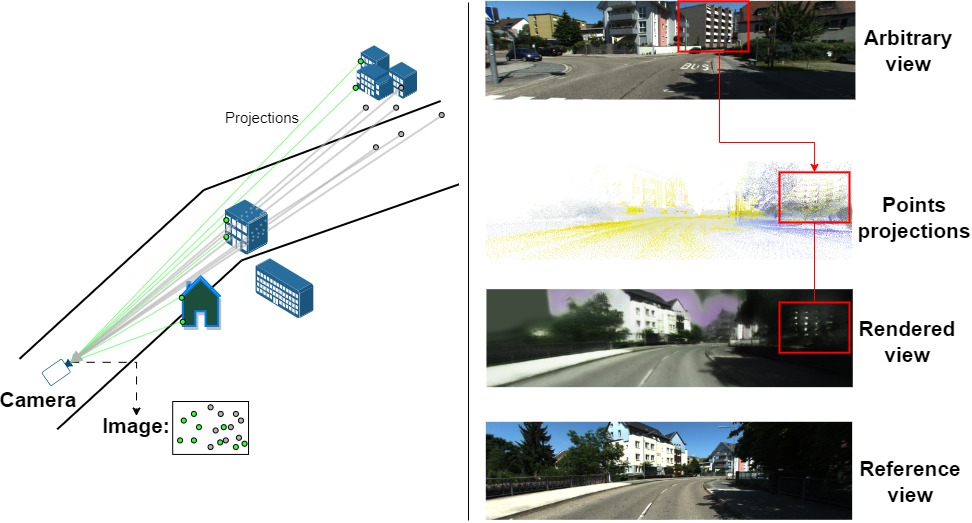}

  \caption{The left side shows the projection of 3D points into the camera, which includes seen (green) and unseen (gray) 3D points. Using these projections to render a view results in artifacts, as shown on the right side. Right: an example showing incoherence between the points projections (geometry) and the reference view (appearance). The resulting renderings are deteriorated. The sample depicts the projection of an unseen building in the current camera view. The building is shown in the arbitrary views.}
  \label{fig:mismatch}
\end{figure}

Rendering photo-realistic scenes has applications in many industries, including autonomous driving, gaming, cinematography, and virtual and augmented reality. Scaling novel view synthesis to large-scale scenes can drive advancements in these fields and potentially reduce costs, but it also presents significant challenges. This is particularly evident when using two different input modalities for rendering: posed images and synchronized LiDAR point clouds. Images provide rich appearance information about the scene but lack geometric data. Conversely, LiDAR point clouds offer reliable geometric information. These two modalities can complement each other and enable several applications, especially novel view synthesis for large-scale scenes.

For fitting scenes and rendering new views, point-based graphics use points as the modeling primitive of the scene \cite{Levoy2000TheUO}. Points and their associated attributes, such as surface orientation, disk radii, and color, are projected into the novel view to create rasterizations from which the view is rendered. Recent deep point-based graphics methods \cite{npbg, npbgpp} aim to encode local photometric and geometric parameters of the surface with neural descriptors, learned during optimization. However, these methods are typically applied to small objects with good coverage, often from 360-degree views.

In this paper, we propose CE-NBPG, a point-based graphics framework for novel view rendering in large-scale autonomous driving scenarios (forward-moving cameras). Our method utilizes a 3D point cloud map of the scene constructed from LiDAR scans and a set of reference images. Each point is associated with a descriptor, learned through a data-driven approach. 

In large-scale autonomous driving scenes, LiDAR scans are commonly accumulated to form a dense point cloud representation of the scene, which is very relevant in the scope of novel view synthesis.
However, the projection of the map point cloud into the image may result in an apparent conflict between what the image sees and what the projected point cloud (geometry) tells about the scene. In this paper, we point out that this results in a discrepancy between the view appearance and the view geometry. As a result, rendering novel views using the 3D point cloud map directly, as in previous methods \cite{npbg, npbgpp}, obtains low and degraded image quality. Fig. \ref{fig:mismatch} illustrates the mismatching between the projected point cloud into the image and the actual view of the image. Projecting unseen points leads to artifacts in the rendered views.

To address the discrepancy between appearance and geometry, we propose establishing a connectivity relationship between the two input modalities, linking the RGB images from a camera with the 3D points from a LiDAR point cloud. Instead of projecting the entire 3D point cloud map into the image, we retrieve and rasterize only the 3D points relevant to the current view based on the connectivity information. This approach not only resolves the conflict between appearance and geometry, leading to improved renderings but also accelerates the fitting and rendering process, making the pipeline more scalable on limited resources. Furthermore, this solution can be constructed once and used off-the-shelf for fitting as well as rendering new views in the given scene. For rendering a view from a novel pose, the 3D points from the closest pose in the connectivity graph are retrieved.

To further improve the quality of renderings, we leverage Generative Adversarial Networks (GANs) and propose a joint adversarial and point rasterization training. We propose to pair a multi-resolution discriminator with a U-Net image synthesizer during training. The image synthesizer processes rasterized neural descriptors to produce RGB images at different resolutions. Meanwhile, the discriminator directs the synthesizer to generate more realistic images and improve the source of neural descriptors. At inference, this pairing is decoupled, i.e., the discriminator is removed, and the synthesizer produces images.

In summary, the contributions of our CE-NBPG framework are the following:
 \begin{itemize}
     \item We identify a new problem which is the incompatibility between appearance and geometry that emerges from large-scale rendering when using the two modalities as input: images and LiDAR point cloud.

     \item We propose to build a connectivity relationship graph between the two modalities, images, and LiDAR point cloud, to tackle the issue of conflict between appearance and geometry for the benefit of novel view fitting and rendering. The connectivity graph is built once and used off-the-shelf for fitting and rendering.

     \item We introduce a joint adversarial and point rasterization training method. This approach incorporates a multi-resolution image-based discriminator with a point-based image synthesizer network to improve neural encoding and rendering quality.

 \end{itemize}
\section{Related Work}
\label{sec:related work}

\noindent\textbf{Novel View Synthesis (NVS):} NVS has been a key research area in computer vision and graphics. Traditional methods used structure-from-motion (SfM) \cite{sfm_classic, sfm} and multi-view stereo (MVS) \cite{mvs_classic, mvs} to re-project and blend images into a new perspective \cite{ibr_chaurasia2013depth, ibr_hedman2016scalable}. With deep learning advancements, newer techniques emerged \cite{ibr_dl_flynn2016deepstereo, ibr_dl_zhou2016view, ibr_dl_kopanas2021point}, including Neural Radiance Fields (NeRFs) \cite{nerf}, which model scenes continuously using ray-marching. Following its introduction, many works were proposed to improve its speed \cite{fastnerf, instantngbnerf}, quality \cite{mipnerf, Nerfusion}, and scalability \cite{blocknerf, meganerf}. Forward-rendering methods, like neural point-based graphics \cite{npbg, npbgpp} and 3D Gaussian splatting \cite{3dgs} offer faster runtimes by using point-based geometry. Our proposal is a forward-rendering approach that works directly on the raw point cloud. Rather than processing the whole point cloud of a scene, it retrieves only relevant scene points and their descriptors and rasterizes them to synthesize the novel view.

\noindent\textbf{Point-based Graphics:} Point clouds have long been favored for scene rendering due to their simplicity and efficiency, offering direct indexability and suitability for forward rendering. Recently, deep learning has been applied to point-based image rendering. Deferred Neural Rendering (DNR) \cite{thies2019deferred} learns neural textures and uses a rendering network to map them to RGB images. Similarly, NPBG \cite{npbg} learns neural descriptors to encode geometry and appearance from raw point clouds, bypassing surface estimation. NPBG++ \cite{npbgpp} improves by estimating view-dependent descriptors with a convolutional feature extractor, reducing fitting time. However, these methods struggle with large-scale scenes. They were designed for small objects and scenes with 360-degree views. Our work addresses this by optimizing neural descriptors for large-scale autonomous driving scenarios.

\noindent\textbf{Large-scale Scenes:} Novel view synthesis for large scenes has been approached using various scene representations. NeRF \cite{nerf} models scenes with a single function, but struggles with large-scale environments. To address this, following methods \cite{blocknerf, meganerf} cluster scenes into parts and assign multiple NeRFs with a retrieval criterion. Neural point-based graphics (e.g., NPBG \cite{npbg} and NPBG++ \cite{npbgpp}) improve runtime but are limited to small objects with sufficient view coverage. These methods compute point visibility by relying on a simple depth buffer which is not suitable for large scenes. READ \cite{read} extends point-based graphics to large-scale scenes but relies on time-consuming, sparse SfM-generated point clouds. Our method uses LiDAR point clouds, common in autonomous driving, and densifies them by accumulating points around the camera in a preprocessing step. In addition, READ addresses point visibility by dividing the world space into regions with N voxels and representing points with 3D cubes. However, this does not address projecting unseen points leading to invalid image regions. Instead of projecting the entire point cloud, we retrieve only visible 3D points using a connectivity relationship between images and points. We also introduce generative adversarial training to enhance the neural point-based graphics approach.

\noindent\textbf{GAN-based rendering:} GANs~\cite{goodfellow2014generative,karras2019style,xie2021style,shen2021tridentadapt}, consisting of a generator and a discriminator, are widely used for data generation and modeling data distribution. Due to their success in vision tasks, researchers have applied GANs to improve image rendering. SinNeRF~\cite{xu2022sinnerf} uses a discriminator for rendering quality but relies on single-scale input, limiting local and global optimization. MPR-GAN~\cite{xu2022mpr} adds multi-scale discriminators but still uses a single-scale generator, hindering progressive refinement. GANeRF~\cite{roessle2023ganerf} refines image quality through multiple downscaled versions but increases computational complexity. Our approach introduces a multi-scale setup for both rendering and the discriminator, enabling progressive refinement at multiple levels with a single synthesizer network acting as the generator. The discriminator is added only for scene fitting.
\section{Method}
\begin{figure*}[h]
\centering
\includegraphics[scale=0.3]{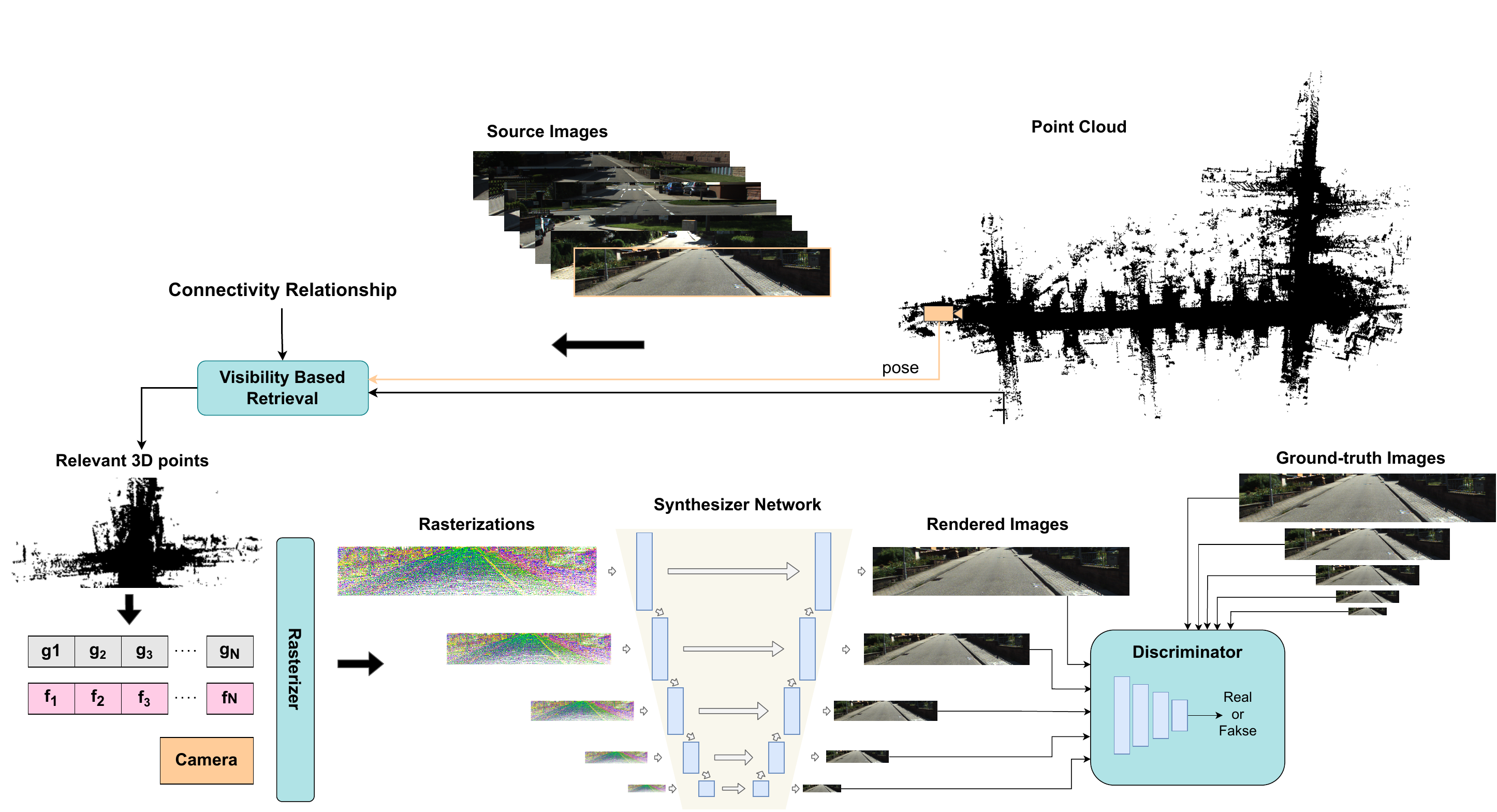}
      \caption{Our CE-NPBG: a scene is represented by source images and a point cloud. To render a novel view (yellow camera), a visibility module retrieves relevant 3D points, whose descriptors are rasterized into the camera at various resolutions. A refiner network then maps the rasterized images into the novel view. During training, a multi-resolution discriminator improves quality by classifying the generated images as real or fake. $\mathbf{f_i}$ is the descriptor for point $\mathbf{g_i}$.}
  \label{fig:sketch}
\end{figure*}

Our CE-NPBG framework, shown in Fig. \ref{fig:sketch}, learns to generate images from novel perspectives using reference images, poses, intrinsic parameters, and LiDAR scans.

Given a large 3D point cloud with $N$ points $G = \{ \mathbf{g}_i, ..., \mathbf{g}_N\}$, each point is associated with an $8$-dimensional neural descriptor $F = \{\mathbf{f}_i, ..., \mathbf{f}_N\}$ that encodes appearance and geometry. For novel view rendering or fitting, our method retrieves visible points from the current camera perspective and projects the 3D points and descriptors into images at various resolutions to form a pyramid of rasterized images. A deep network, the synthesizer, then converts these rasterized images into an RGB image. We update the synthesizer's weights and optimize neural descriptors using an adversarial loss and reconstruction loss, minimizing the difference between predicted and reference images. Each module is described in detail below.

\begin{figure*}[ht!]
\scriptsize
  \centering
  \setlength{\tabcolsep}{2.0pt}
  \begin{tabularx}{\textwidth}{ c c c c}

    
    \includegraphics[width=0.24\textwidth]{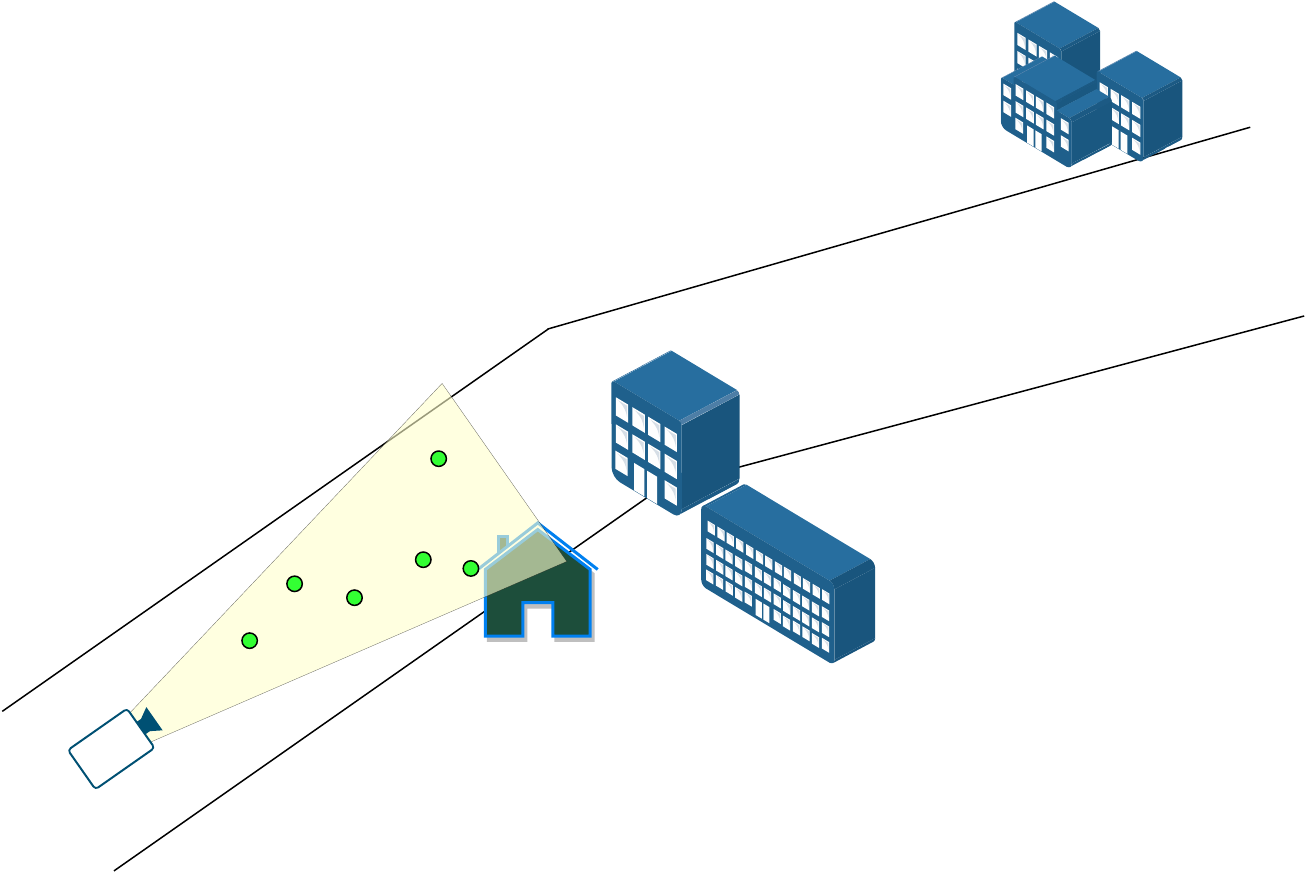}
     &
     
    \includegraphics[width=0.24\textwidth]{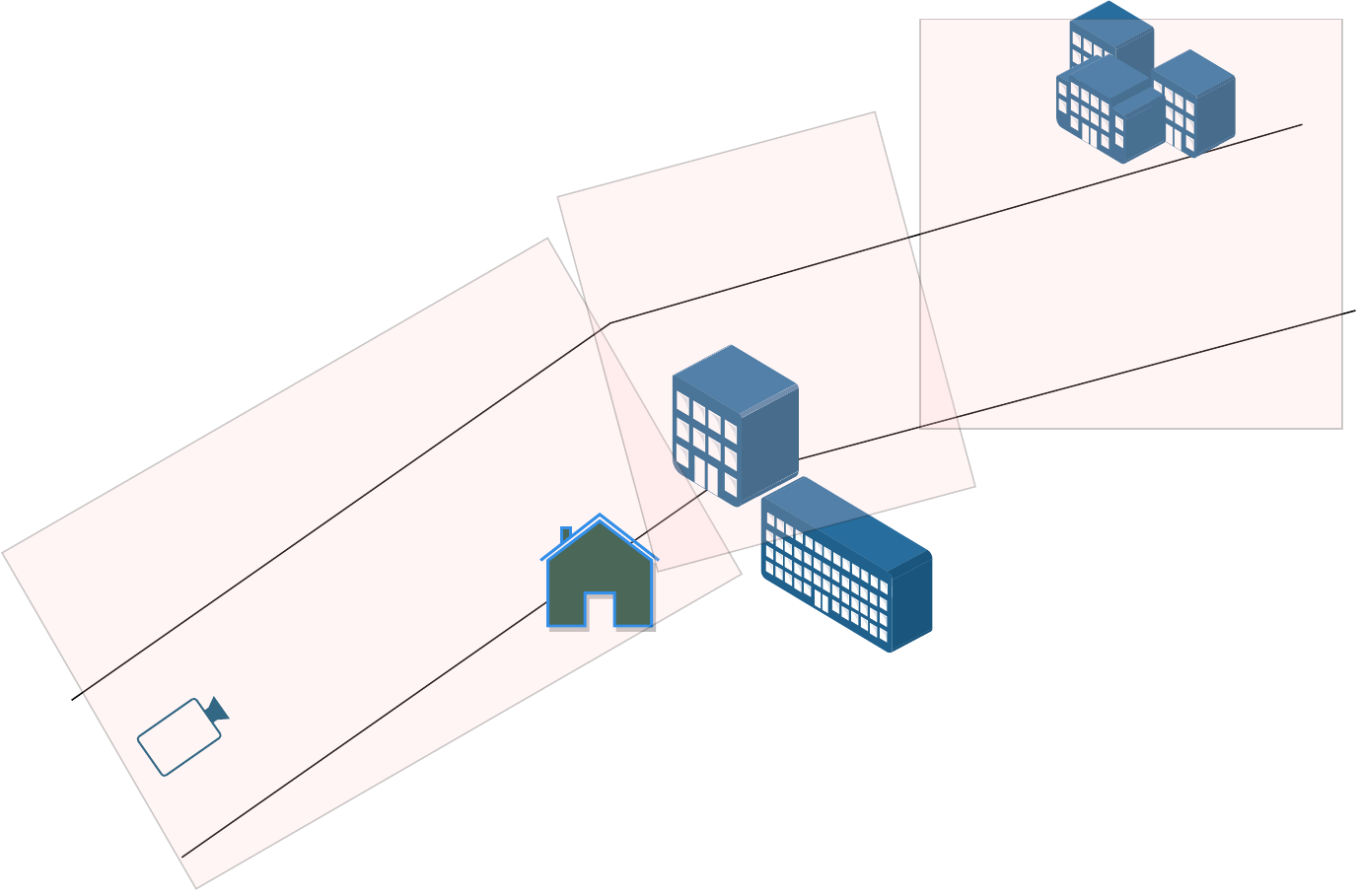}
    
     &
     
    \includegraphics[width=0.24\textwidth]{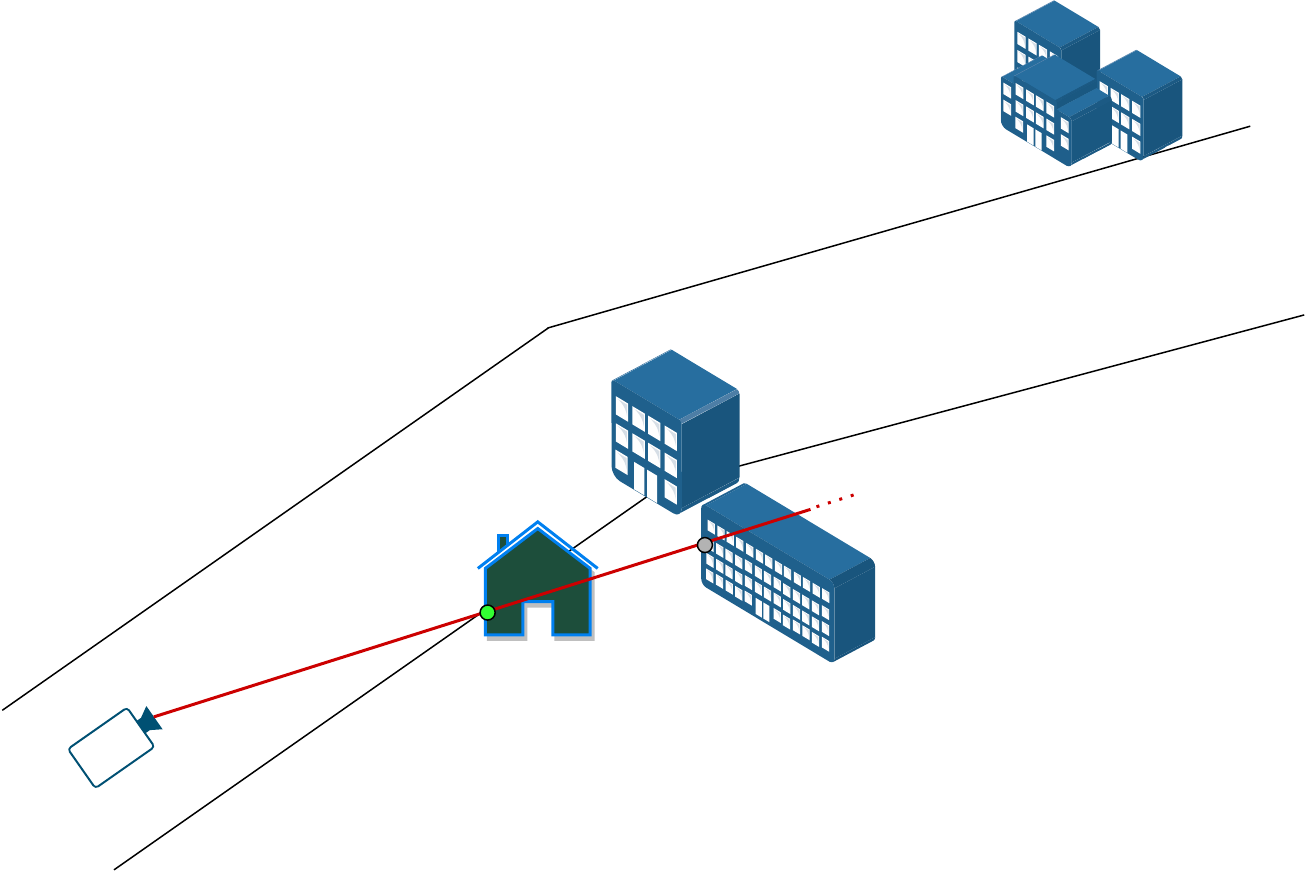}
    &

    \includegraphics[width=0.24\textwidth]{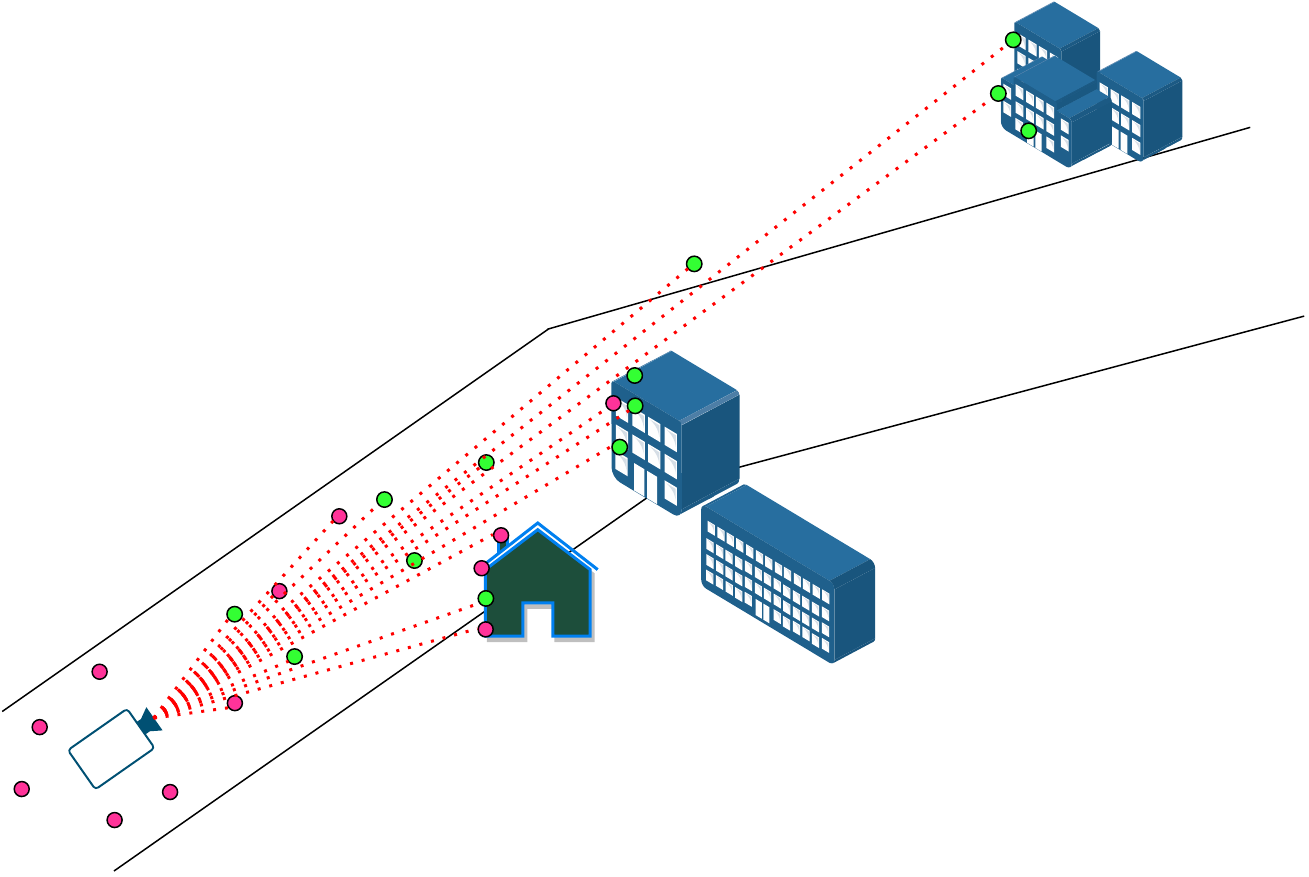}

    \\ 

    a & b & c &d \\
    
  \end{tabularx}
  \caption{Different proposals to address large point cloud rendering: a) clipping points outside certain threshold, b) clustering of the scene, c) taking closest points based on depth-buffer d) Our CE-NPBG: building connectivity relationship between 3D points and poses/images.} \label{visibility}
\end{figure*}

\noindent\textbf{Visibility Estimation from Connectivity Relationship:} \label{connectivity}
Some 3D points are not visible from a camera's perspective due to occlusions but they can still be projected onto the image plane, resulting in a mismatch between the visual appearance from that camera perspective and the geometry derived from the 3D point cloud map (Fig. \ref{fig:mismatch}). To address this issue, several solutions can be proposed (Fig. \ref{visibility}). (i) One approach is to filter out points with depth values exceeding a certain threshold (Fig. \ref{visibility} a), but this is impractical for large-scale environments, as the threshold must constantly adapt to the camera's range of view. Additionally, in many cases, the image will capture more than the selected 3D points convey about the environment. (ii) The same issue arises with sliding windows, an extension of thresholding, which includes points behind the camera. (iii) Clustering the scene based on 3D points or splitting it into blocks is another option (Fig. \ref{visibility} b), but it introduces complexities like overlapping boundaries and the need to train different networks for each division. (iv) Another way is to select the closest point via a depth-buffer (Fig. \ref{visibility} c) \cite{npbg, npbgpp}, but this requires dense pixel coverage, which is impractical for larger scenes. (v) Another solution is to learn visibility prediction, though this adds network complexity and may not generalize effectively. In contrast to these solutions, we build a connectivity relationship between the posed RGB images and the 3D points from the point cloud maps (Fig. \ref{visibility} d). This relationship is not naturally available, as the two entities come from different sensor modalities.

We establish the connectivity relationship in two stages: a greedy stage followed by a pruning stage. First, we aggregate LiDAR scans into a single point cloud, creating a denser scene representation. In the greedy stage, 3D points are grouped with corresponding images while building connectivity. For an image at timestamp \textit{t}, we use several nearby LiDAR scans (local isolation) to ensure overlap between the image and the current section of the scene. Specifically, we take $n$ scans behind and 2$\times n$ scans ahead of the camera, ensuring denser point coverage. This stage assumes all points in the isolated section are visible from the image's pose.

After building this generous connectivity graph, we apply pruning to account for local occlusions. For a given camera pose (during scene fitting or rendering), our method looks in the connectivity relationship for the closest camera pose and retrieves all of its connected 3D points. The pruning stage follows, in which only points that fall within the boundaries of the image and pass the depth checks are kept, and the remaining points are dropped. The depth checks include points that lie ahead of the camera (positive depth) and are the closest to the camera (shortest depth if multiple points fall onto the same pixel). These are the visible points. 

During scene fitting, only the descriptors of the retrieved points are updated in each iteration.
To sum up, this connectivity graph is built once and can be used (off-the-shelf) for training and testing (rendering). The building stage applies local isolation, detaching the local scene from the whole scene and guaranteeing that far scene points that can project into the image (but not observed) are not visible by the current view. The pruning stage applies visibility checks to prune off occluded and locally unobserved points.

\noindent\textbf{Rasterization:}
After retrieving the visible points from the camera perspective, our system projects the 3D points and associated descriptors into a series of down-sampled resolutions, forming a pyramid of rasterized raw images ${\textbf{S}}_{t=1}^{T}$, where each raw image $\textbf{S}_t \in \mathbb{R}^{\frac{H}{2^t} \times \frac{W}{2^t} \times C}$ (with $T=5$ in our experiments). These raw images provide the initial appearance information for the synthesizer to map into RGB values. The lowest resolution gets densely filled and thus suffers from fewer holes and bleeding, while the highest captures fine details. Our connectivity ensures that only visible descriptors are rasterized, avoiding updates to occluded points during scene fitting.

\noindent\textbf{Image Synthesis:}
The synthesizer network, a U-Net with gated convolutions, processes rasterized images to generate a full-resolution RGB image in one pass. It fuses coarse and fine details from the input pyramid to cover bleeding surfaces. In addition to producing the image at the highest resolution, the network also renders views at down-sampled resolutions, designed to pair with the multi-scale discriminator, which is explained next.

\noindent\textbf{Joint Adversarial-rasterization training:} \label{joint_training}
Apart from RGB regression, we want the synthesizer network to learn generative capabilities to produce realistic outputs with fewer artifacts, even when point cloud information is insufficient. This also helps the synthesizer generalize better on unseen inputs. To improve rendering quality, we integrate concepts from generative adversarial networks (GANs) into the point-based synthesizer via joint adversarial training.

We append the synthesizer network $R_\theta$ with a discriminator $D_\phi$, using the synthesizer as the generator. To avoid adding extra complexity, the discriminator is used only during training and removed at test time. We enhance the discriminator with multi-scale branches for different image resolutions. During training, the discriminator distinguishes between real (ground-truth) and fake (generated) images, while the synthesizer aims to fool it, improving output quality. Following LSGAN~\cite{mao2017least} criteria, the adversarial losses are formulated as:
\begin{align}
    \label{eq:adv_refiner}
    &{L^{R_\theta}_{adv} =\sum^{N_{s}}_{i=1} (D_\phi^{(i)}( R_\theta^{(i)}(\bigcup_{t=1}^T \mathbf{S}_{t})) -1)^{2} }
\end{align}
\begin{align}
    \label{eq:adv_disc}
    &{L^{D_\phi}_{adv} =\sum^{N_{s}}_{i=1} [D_\phi^{(i)}( R_\theta^{(i)}(\bigcup_{t=1}^T \mathbf{S}_{t}))^{2}} +(D_\phi^{(i)}(\mathbf{I}_{gt}^{(i)}) -1)^{2}] 
\end{align}

where ${i}$ is the scale index and ${N}_{s}$ denotes the total number of discriminator scales and $\mathbf{I}_{gt}^{(i)}$ is the scale-adjusted ground-truth image. In our experiments, we use ${N}_{s}$ of 5.

Besides the adversarial loss, we use the perceptual loss \cite{johnson2016perceptual} that computes the difference between the activations of a pretrained VGG network using the rendered and ground-truth images.

\section{Experiment}
\label{sec:experiment}

\subsection{Datasets} \label{data_sets_describe}
We use scenes from the KITTI360 dataset, a large, real-world driving dataset designed for tasks like novel view synthesis. KITTI360 improves upon the original KITTI dataset with denser scene coverage and more accurate, geolocalized vehicle and camera poses. We select sequences from three different trajectories, dividing each into training and testing parts. Every $10^{th}$ frame is used for testing, with the remaining frames for training.

We build the 3D point cloud map of each scene by accumulating the 3D points from the LiDAR scans. This delivers a dense and compact 3D point cloud. In our experiments, we use a single GPU
with a capacity of 48 Gigabytes. Accordingly, we tune our selection of the sequence data size (images and 3D points) to fit the given GPU space. Given that, we can select a sub-sequence with slightly more than 300 frames and a number of 3D points close to 37 million. These are summarized in Tab. \ref{kitti_description}.

\begingroup
\setlength{\tabcolsep}{1.0pt}
\begin{table}[h]
\caption{Details of the sub-sequences from KITTI360 that are used in our experiments. The indices (0, 4, and 6) correspond to the original sequence number of KITTI360.}
\begin{center}

\begin{tabular}{@{}lccc@{}}\toprule

&  KITTI-0 &  KITTI-4 &  KITTI-6  \\

 \cmidrule{2-4}

Frames indices & 9602 $\rightarrow$ 9918 & 9975 $\rightarrow$ 10275 & 8796 $\rightarrow$ 9106 \\
Number of frames & 316 & 300 & 310\\

Number of points & 36950232 & 36352862 & 37376839\\

\hline
\end{tabular}
\end{center}
\label{kitti_description}
\end{table}
\endgroup

\begin{table*}[ht]
\caption{Comparison of our method against the state-of-the-art methods on sub-sequences from the KITTI360 dataset.}
\label{ours_vs_sota}
\begin{center}
\scriptsize

\begin{tabular}{@{}rcccccccccccc@{}}\toprule

&  & \multicolumn{3}{c}{\textbf{KITTI-0}} & \phantom{} & \multicolumn{3}{c}{\textbf{KITTI-4}} & \phantom{} & \multicolumn{3}{c}{\textbf{KITTI-6}} \\

Method &  & PSNR $\uparrow$ & SSIM $\uparrow$ & LPIPS $\downarrow$ & & PSNR $\uparrow$ & SSIM $\uparrow$ & LPIPS $\downarrow$ & & PSNR $\uparrow$ & SSIM $\uparrow$ & LPIPS $\downarrow$\\

\cmidrule{3-5} \cmidrule{7-9} \cmidrule{11-13}

NPBG \cite{npbg} & & 21.59 & 0.60 & 0.38  & & 23.43 & 0.61 & 0.39 & & 22.89  & 0.61 & 0.38 \\

NPBG++ \cite{npbgpp} & & 20.92 & 0.71 & \textbf{0.32} & & 22.91& 0.72 & 0.33 & & 23.61 & 0.75& 0.29\\

READ \cite{read} & & 14.41& 0.49 & 0.49 &  & 16.57& 0.52& 0.48 &  & 15.30& 0.53& 0.46\\

DS-NeRF \cite{dsnerf} & & 15.05& 0.46 & 0.62 &  & 17.67& 0.52& 0.60 &  & 16.81& 0.52& 0.56\\

3DGS \cite{3dgs} & & 16.99& 0.48 & 0.56 &  & 18.15& 0.50& 0.55 &  & 17.07& 0.51& 0.55\\

\cmidrule{1-13}

CE-NPBG (Ours) & & \textbf{22.49}  & \textbf{0.77} & 0.34 & & \textbf{24.33} & \textbf{0.75} & \textbf{0.33} & & \textbf{24.17}& \textbf{0.79}& \textbf{0.28} \\

\textit{Ours w/o adversarial training} & & \textit{22.26}  & \textit{0.75} & \textit{0.35} & & \textit{24.11} & \textit{0.74} & \textit{0.34} & & \textit{23.82}& \textit{0.77}& \textit{0.29} \\

\bottomrule
\end{tabular}
\end{center}
\end{table*}

\subsection{Baselines Methods}
We compare our method to three state-of-the-art neural point-based rendering methods, a NerF-based approach, and 3D Gaussian splatting. All methods use the same data sources: LiDAR 3D point clouds and posed reference images. NPBG \cite{npbg} and NPBG++ \cite{npbgpp} augment each point with a learnable descriptor, using depth-buffer for visibility, while NPBG++ adds multiview optimization. READ \cite{read} handles visibility by voxelizing the scene and representing points with 3D cubes. For Nerf, we use DS-NeRF \cite{dsnerf}, which utilizes depth (we obtain it from LiDAR) as an additional supervision signal. 3D Gaussian splatting 3DGS \cite{3dgs} optimizes Gaussian parameters to render novel views.

\subsection{Training Details}
We conduct the experiments on the baselines using the same source of 3D data: accumulated LiDAR point clouds. We set the number of LiDAR scans $n$ to 5 for connectivity relationship building.
All of the point-based baselines use a U-Net-like architecture for regressing the novel view.
For DS-NeRF training, we had to do certain adaptations to obtain the rendering. We transformed the global coordinate frame of the whole sequence to the first frame of the local sequence (Sec. \ref{data_sets_describe}) and transformed the poses of the images into this local coordinate frame. Besides, we normalized the camera and 3D points positions. For 3DGS, we had to cut further the sub-sequences because of its huge memory requirements (further on this in Sec. \ref{ours_for_3DGS}). For its optimization on KITTI scenes, we minimized the position and scaling learning rates.

\subsection{Comparison with Prior Arts}
Table \ref{ours_vs_sota} presents quantitative results using the standard rendering metrics: Peak Signal-to-Noise Ratio (PSNR), Structural Similarity (SSIM), and learned Perceptual Image Patch Similarity (LPIPS). As can be observed, our method exceeds the different state-of-the-art rendering methods. The main reason for that is our visibility solution, which addresses the incompatibility between appearance and geometry. It uses the visible points to render novel views.
Previous point-based methods estimate point visibility using different approaches, which show limitations in large-scale scenes. They compute a depth buffer projecting the point cloud into the image and taking the nearest points as the visible points. NPBG++ \cite{npbgpp}, in addition, rasterizes the point cloud into an image with reduced size, where the size is determined a priori by setting a visibility reduction factor to account for the sparsity of a point cloud. While taking the closest point can mitigate the impact of occluded parts, it is unreliable enough to address large point cloud maps and prevent the influence from unseen points. READ further implements a cube rasterization scheme to account for invalid regions in the image plane. Rather than projecting the 3D points themselves of a point cloud directly onto the image plane as in NPBG and NPBG++, it adapts the rasterization scheme by representing a 3D point with a cube. It divides the world space into regions with N voxels and projects the cubes instead of points into the image plane to reduce the impact of holes in the sparse point cloud and the resulting invalid regions in the image plane. 3D Gaussian Splatting (3DGS) shows limitations in addressing large scale scenes. It blends a set of depth-ordered points which include many unseen points to render a view. In the next section, we show the benefit of our connectivity relationship to 3DGS.

Rather than rasterizing the whole 3D point cloud map into the image, we rasterize the visible ones. Our connectivity relationship shows its advantage in determining the visible points in the current frame and retrieving them. Our system, as a result, makes the geometry that is projected into the image and the appearance of the image compatible, which, in consequence, improves the rendering quality.

DS-NerF fails as well to scale to large scenes, showing the limitation of the single multi-layer perceptron to address large scale scenes. Nerf based methods, even when supervised with dense depth signals, cannot scale up to large scenes. For scalability, the scene has to be clustered and addressed with different Nerfs. Such volumetric methods excel in small and bounded scenes where there is a dense view coverage of the scene.

\begin{figure*}[ht!]
\scriptsize
  \centering
  \setlength{\tabcolsep}{2.0pt}
  \begin{tabularx}{\textwidth}{r c  c  c }


    Reference
    
    &    
    \includegraphics[width=0.28\textwidth]{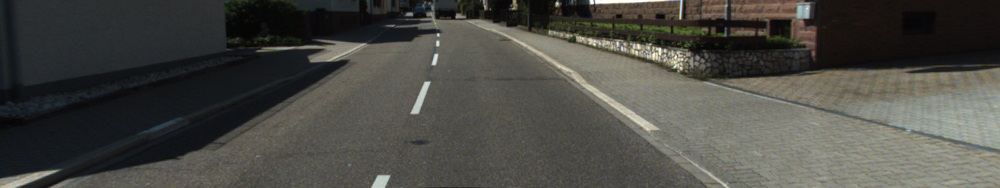}
    &
    \includegraphics[width=0.28\textwidth]{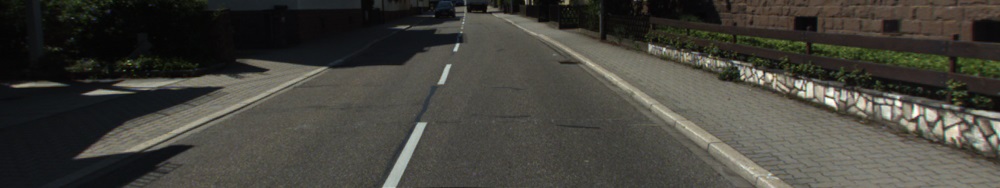}
    &
    \includegraphics[width=0.28\textwidth]{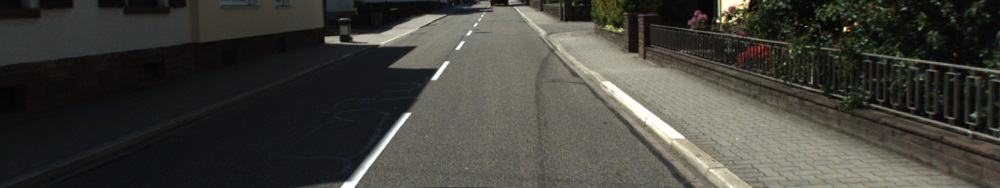}
    \\ 
    
     Ours
    &    
    \includegraphics[width=0.28\textwidth]{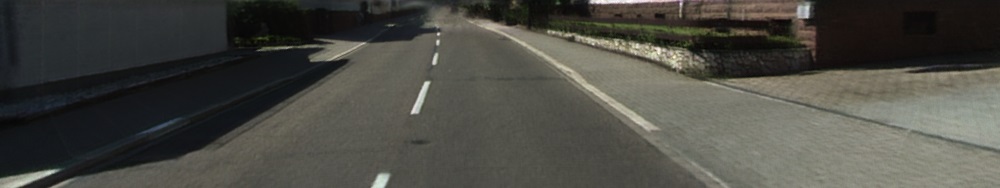}
    &
    \includegraphics[width=0.28\textwidth]{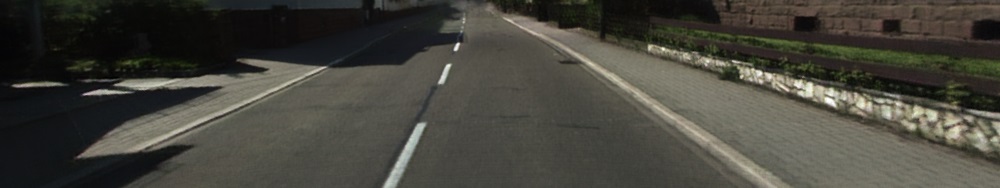}
     &
    \includegraphics[width=0.28\textwidth]{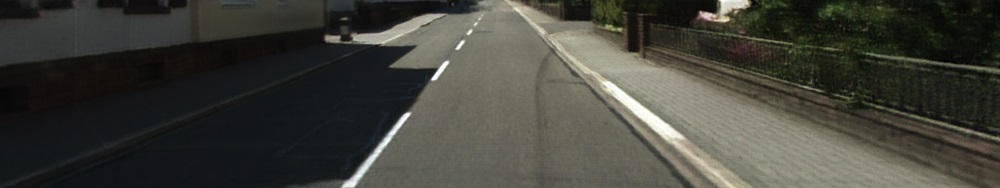}
    \\ 

     NPBG
    &    
    \includegraphics[width=0.28\textwidth]{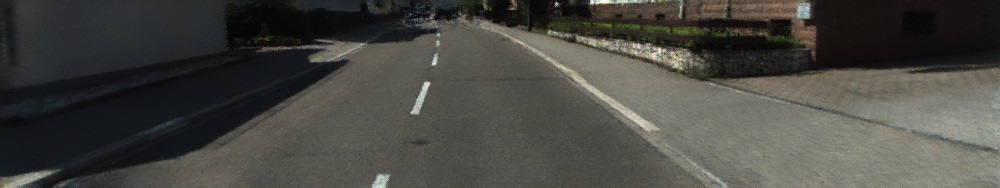}
    &
    \includegraphics[width=0.28\textwidth]{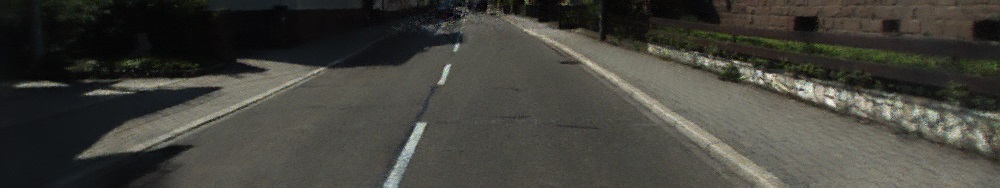}
    &
     \includegraphics[width=0.28\textwidth]{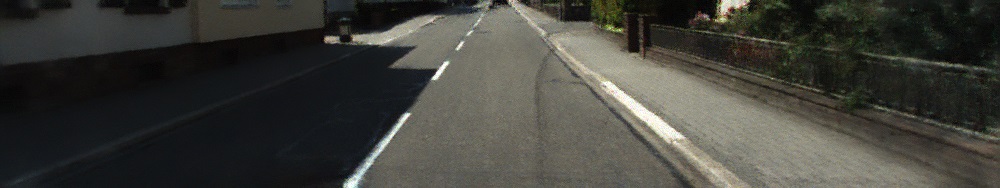}
    \\ 

    NPBG++
    &    
    \includegraphics[width=0.28\textwidth]{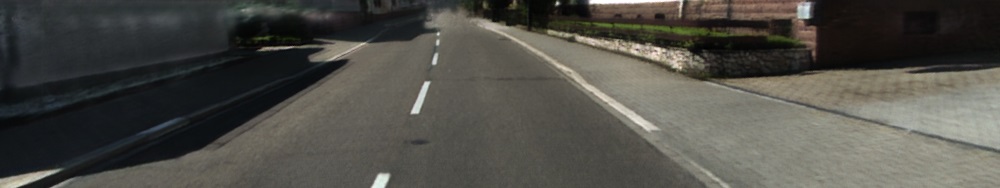}
    &
    \includegraphics[width=0.28\textwidth]{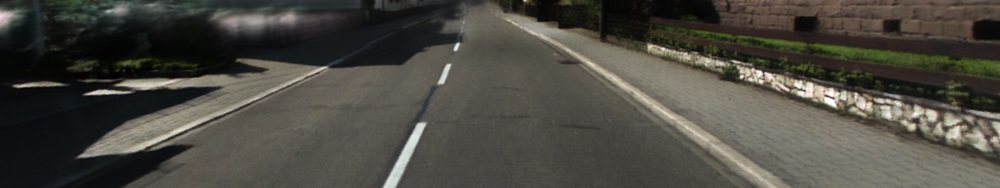}
     &
     \includegraphics[width=0.28\textwidth]{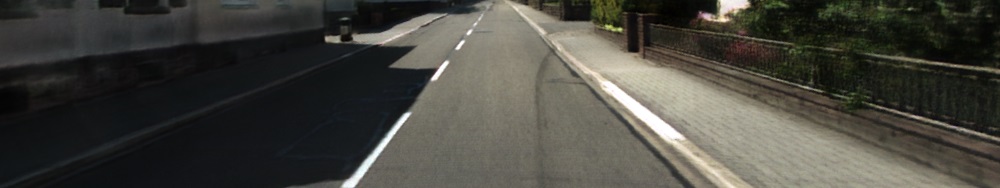}
    \\ 

    READ &
     \includegraphics[width=0.28\textwidth]{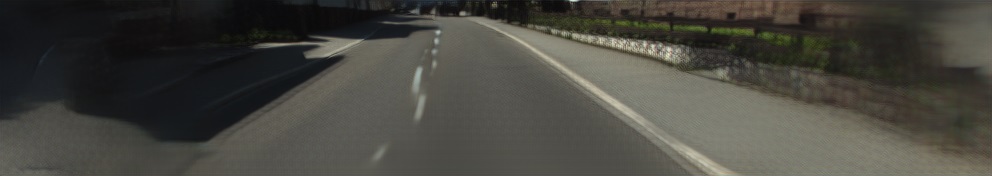}
    &    
    \includegraphics[width=0.28\textwidth]{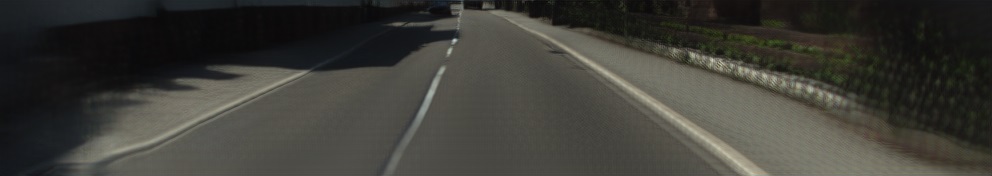}
    &
    \includegraphics[width=0.28\textwidth]{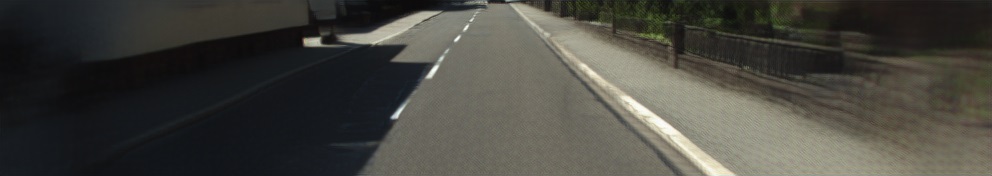}

    \\ 

    DS-NeRF &
     \includegraphics[width=0.28\textwidth]{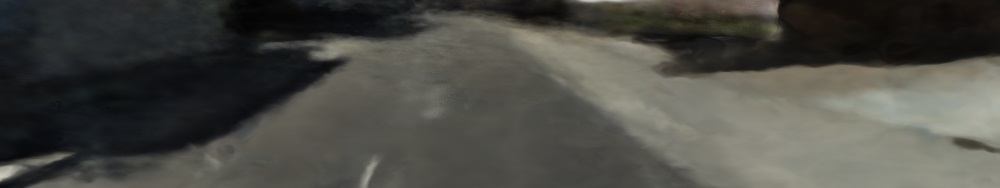}
    &    
    \includegraphics[width=0.28\textwidth]{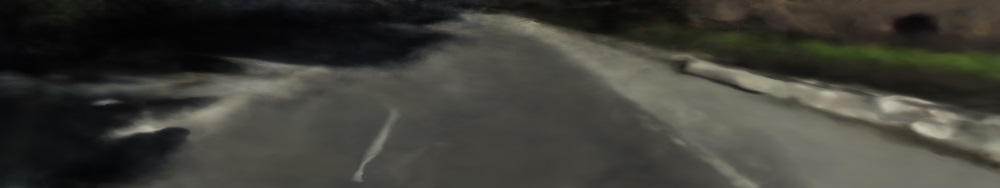}
    &
    \includegraphics[width=0.28\textwidth]{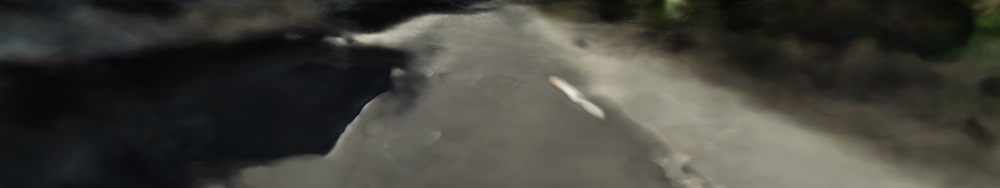}

    \\ 

    3DGS &
     \includegraphics[width=0.28\textwidth]{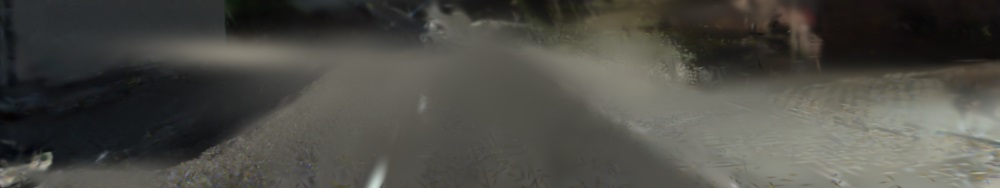}
    &    
    \includegraphics[width=0.28\textwidth]{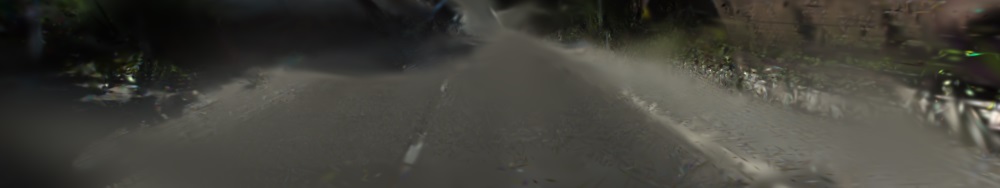}
    &
    \includegraphics[width=0.28\textwidth]{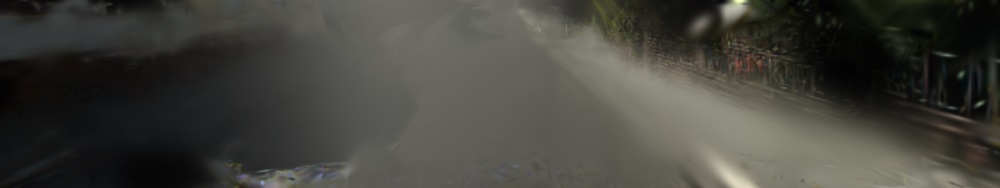}

    \\ 
    
  \end{tabularx}
  \caption{Rendering results of our method and other state-of-the-art methods.}\label{qualitative_ours_vs_sota}
\end{figure*}

In Fig. \ref{qualitative_ours_vs_sota}, we complement these results with qualitative results by showing rendering results from novel views. As can be observed, the renderings from our method are the ones closer to the reference images. NPBG++ \cite{npbgpp}, for example, produces white artifacts and unclear scenery. This is mainly due to the limitation of its visibility estimation approach.

Large unbounded autonomous driving scenes pose difficulties to existing methods. Our method, however, succeeds in addressing large scenes challenges such as sparse view coverage and the appearance-geometry mismatch. 

\subsection{Usability and scalability of our method} \label{ours_for_3DGS}
In the previous experiment, we have shown the advantage of our method in addressing large scale autonomous driving scenes, compared to neural point-based and other rendering methods. Practically, we have selected 3D points and images to fit the maximum capacity of the GPU. Further scalability boils down to scaling up the available resources.
In this experiment, we show the benefit of our visibility solution for the usability and scalability of other forward-rendering (rasterization-based) approaches such as the 3D Gaussian Splatting (3DGS) \cite{3dgs}.

\begin{table}[ht]
\caption{Utilizing our connectivity relationship graph for 3D Gaussian Splatting (3DGS) \cite{3dgs} scene fitting.}
\label{gsn_connectivity}
\scriptsize
\setlength{\tabcolsep}{5.0pt}
\begin{tabular}{@{}lcccccc@{}}\toprule

&  \multicolumn{3}{c}{\textbf{KITTI-4-reduced}} & \multicolumn{3}{c}{\textbf{KITTI-4}} \\

Method &  PSNR $\uparrow$ & SSIM $\uparrow$ & LPIPS $\downarrow$ &  PSNR $\uparrow$ & SSIM $\uparrow$ & LPIPS $\downarrow$ \\
\cmidrule{2-4} \cmidrule{5-7}

3DGS \cite{3dgs} &   19.61 & 0.58 & 0.57  &  \multicolumn{3}{c}{fail} \\
\cmidrule{2-4} \cmidrule{5-7}
3DGS + Our &  \multirow{2}{*}{\textbf{28.57}}& \multirow{2}{*}{\textbf{0.83}} & \multirow{2}{*}{\textbf{0.23}} &  \multirow{2}{*}{\textbf{28.65}}& \multirow{2}{*}{\textbf{0.83}}& \multirow{2}{*}{\textbf{0.24}} \\

Connectivity & & & & & &   \\

\bottomrule
\end{tabular}
\end{table}

\noindent\textbf{Usability of connectivity relationship for 3DGS fitting:}
For the purpose of this section, we show the results of the experiments on one sub-sequence (KITTI4). In the previous section, we had to run 3DGS on a subset of the KITTI sub-sequences because the available resources (48 GB) were not enough to optimize the Gaussians for the whole sub-sequences. To account for the 3DGS memory constraints, we show the results as well on the subset of KITTI4, which is KITTI-4-reduced with a total number of 3D points of 19.4 million.
We apply our connectivity-based visibility to 3DGS \cite{3dgs} and evaluate scene fitting performance (i.e., evaluation of the renderings from reference camera poses). 
3DGS \cite{3dgs} proposes to densify a point cloud with additional Gaussians to cover gaps in the reconstruction. Since the accumulated LiDAR point cloud is relatively dense, we deactivate the densification option. Tab. \ref{gsn_connectivity} reports the quantitative results, with and without our connectivity-based visibility solution.

\begin{figure}[ht!]
\scriptsize
  \centering
  \setlength{\tabcolsep}{2.0pt}
  \begin{tabularx}{\textwidth}{ c  c  c }

    Reference & 3DGS & 3DGS + Our Connectivity\\ 
    
    \includegraphics[width=0.16\textwidth]{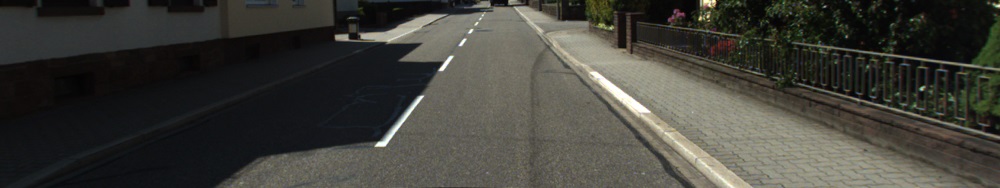}
    &    
    \includegraphics[width=0.16\textwidth]{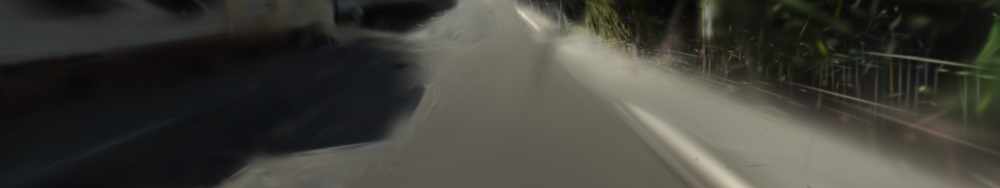}
    &
    \includegraphics[width=0.16\textwidth]{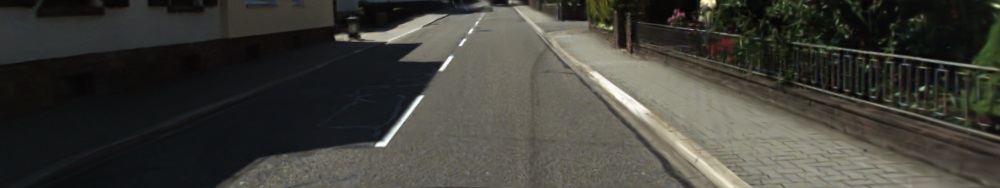}

    \\ 
    
    \includegraphics[width=0.16\textwidth]{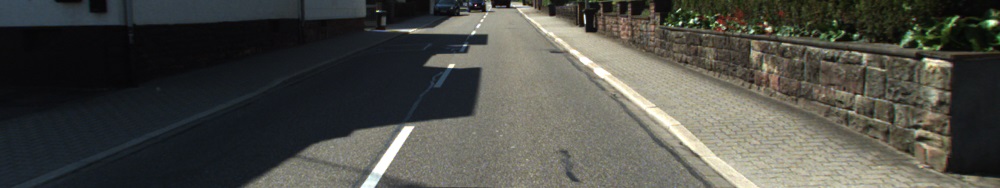}
    &    
    \includegraphics[width=0.16\textwidth]{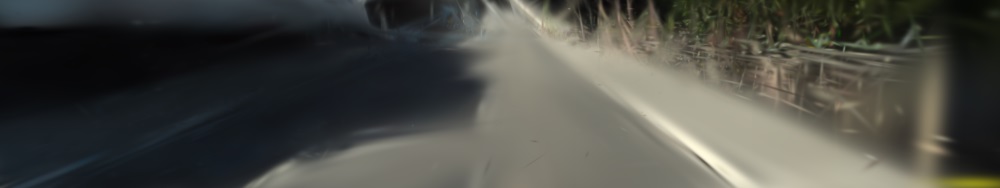}
    &
    \includegraphics[width=0.16\textwidth]{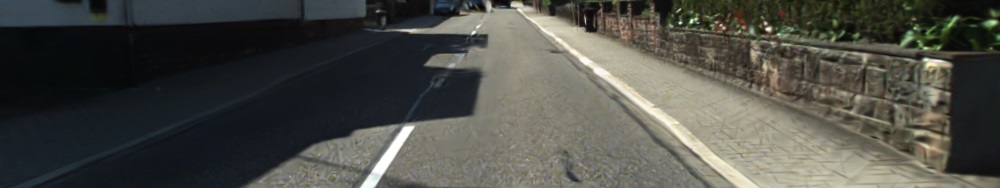}

    \\ 

    \includegraphics[width=0.16\textwidth]{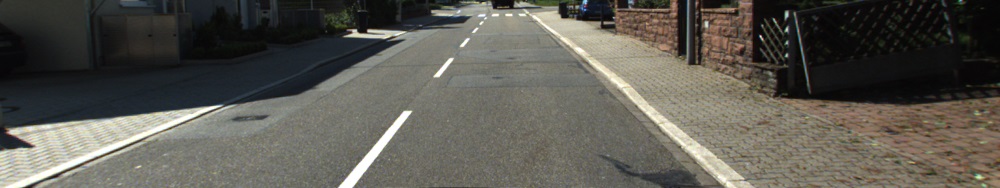}
    &    
    \includegraphics[width=0.16\textwidth]{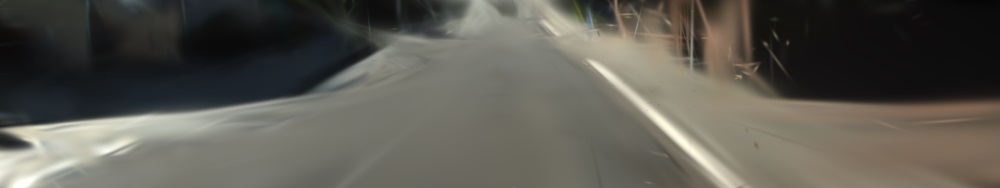}
    &
    \includegraphics[width=0.16\textwidth]{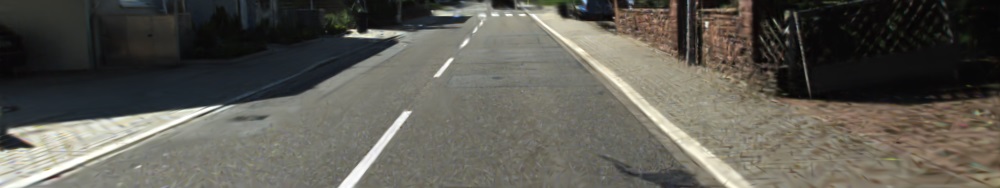}

    \\ 
    
  \end{tabularx}
  \caption{Using our proposed connectivity relationship for visibility estimation drastically improves scene reconstruction using 3D Gaussian Splatting \cite{3dgs}.}
  \label{3dgs_with_without_connec}
\end{figure}

The reported results in the second column of Tab. \ref{gsn_connectivity} (KITTI-4-reduced) show a considerable improvement in the quality of scene fitting when using our connectivity setup.
In the 3DGS setup (without our connectivity relationship), a set of depth-ordered points overlapping the pixel, among them many unseen points, are blended to obtain the color. Whereas, by using our connectivity setup, only the retrieved visible points are blended to compute the color, which results in better optimization. We accompany this analysis with qualitative results in Fig. \ref{3dgs_with_without_connec}. As can be observed, our visibility solutions enhance 3DGS scene fitting for large scenes.

\noindent\textbf{Connectivity relationship for scalability:}
3DGS optimizes not only the 3D points but also the associated Gaussian attributes, which require much additional memory.
The third column of Tab. \ref{gsn_connectivity} presents the results of 3DGS scene fitting with and without our connectivity solution on the whole KITTI-4 sub-sequence which possesses a number of points (approx. 36.3 million points), that is almost the double of KITTI-4-reduced.
Thus, we optimize 3DGS on the whole KITTI-4 sub-sequence (approx. 36.3 million points), which is almost double the size of KITTI-4-reduced (approx. 19.4 million points).
Executing 3DGS on the KITTI-4 sequence ended in a failed reconstruction due to the needed additional memory for optimization. However, the optimization was completed successfully (on the same GPU memory resources) by applying our connectivity relationship (3DGS + Our Connectivity). Our connectivity relationship retrieves only visible points, a small subset of the whole point cloud. This reduces computation and memory requirements, allowing for optimization of large scenes.

Our further NVS experiments have shown that 3DGS \cite{3dgs} still underperforms our neural point-based rendering approach, even when combined with our connectivity relationship. We attribute 3DGS's limitations to the sparse scene coverage in the KITTI360 dataset, where the camera moves forward at car speed.

\noindent\textbf{Connectivity relationship for run-time:}
We report further the run-time improvement from using our connectivity relationship. With the connectivity relationship, visible points are estimated and used for fitting and rendering. This, as a result, accelerates the rendering function by more than 50 times as shown in Tab. \ref{run-time_3dgs}
\begin{table}[ht]
\caption{Run-time improvements from using our connectivity relationship on 3DGS.}
\label{run-time_3dgs}
\begin{center}
\setlength{\tabcolsep}{0.1pt}
\small
\begin{tabular}{@{}lcc@{}}\toprule

Method & Number of points & Render time (s) / FPS \\
\cmidrule{2-3}

3DGS &  19403162 & 0.046 / 21.6 \\
3DGS +Our Connectivity & 720000 & 0.00089 / 1114 \\ 

\hline
\end{tabular}
\end{center}
\end{table}

\subsection{Ablation study: joint adversarial training}
The last row of Tab. \ref{ours_vs_sota} reports the results of our method without the inclusion of discriminator for scene fitting. Comparing these to our methods' results (with discriminator) shows the benefit of the proposed joint point-based rasterization and adversarial training as illustrated in Sec. \ref{joint_training}.
\section{Conclusion}
\label{sec:conclusion}
We present CE-NPBG, a novel neural point-based rendering solution for large-scale autonomous driving scenes that learns neural encodings of 3D points using LiDAR point clouds and reference views. Our system, based on three key contributions, first identifies the main cause of artifacts when combining LiDAR and camera data—mismatches between reference images and rasterized geometry. Second, We align these data sources by building a connectivity graph, rendering only visible points to improve efficiency, demonstrated with 3D Gaussian Splatting. Third, We enhance neural encodings and rendering quality using an adversarial generator-discriminator setup. Future work will explore incorporating language embeddings and semantics for text-conditioned novel view synthesis and scene manipulation.
{
    \small
    \bibliographystyle{ieeenat_fullname}
    \bibliography{main}
}


\end{document}